\documentclass[10pt,twocolumn,letterpaper]{article}
\usepackage{cvpr}      % To produce the REVIEW version
\usepackage{graphicx}
\usepackage{amsmath}
\usepackage{amssymb}
\usepackage{booktabs}
\usepackage{xcolor}
\usepackage{longtable} % Required for multi-page tables
\usepackage{lipsum} % For generating dummy text
\usepackage{xtab} % Required for long tables that span both columns
\usepackage{lipsum} % For generating dummy text
\usepackage{multicol} % Required for multi-column sections
\usepackage{booktabs}  % For professional rules
\usepackage{multirow}  % For grouping rows
\usepackage{array}     % For better column formatting

\usepackage[pagebackref,breaklinks,colorlinks]{hyperref}

\usepackage[capitalize]{cleveref}
\crefname{section}{Sec.}{Secs.}
\Crefname{section}{Section}{Sections}
\Crefname{table}{Table}{Tables}
\crefname{table}{Tab.}{Tabs.}

\def\cvprPaperID{*****} % *** Enter the CVPR Paper ID here
\def\confName{CVPR}
\def\confYear{2025}

\begin{document}

\title{Bridging Robustness and Efficiency: Real-Time Low-Light Enhancement via Attention U-Net GAN}

\author{
  \textbf{Yash Thesia} \\
  Dept. of Computer Science \\
  New York University \\
  New York, NY, USA \\ % Added City, Country
  \texttt{yt2188@nyu.edu}
\and
  \textbf{Meera Suthar} \\
  Dept. of Computer Science \\
  New York University \\
  New York, NY, USA \\ % Added City, Country
  \texttt{ms12418@nyu.edu}
}
\maketitle

\begin{abstract}
Recent advancements in Low-Light Image Enhancement (LLIE) have focused heavily on Diffusion Probabilistic Models, which achieve high perceptual quality but suffer from significant computational latency (often exceeding 2-4 seconds per image). Conversely, traditional CNN-based baselines offer real-time inference but struggle with "over-smoothing," failing to recover fine structural details in extreme low-light conditions. This creates a practical gap in the literature: the lack of a model that provides generative-level texture recovery at edge-deployable speeds.

In this paper, we address this trade-off by proposing a hybrid \textbf{Attention U-Net GAN}. We demonstrate that the heavy iterative sampling of diffusion models is not strictly necessary for texture recovery. Instead, by integrating Attention Gates into a lightweight U-Net backbone and training within a conditional adversarial framework, we can approximate the high-frequency fidelity of generative models in a single forward pass. Extensive experiments on the SID dataset show that our method achieves a \textbf{best-in-class LPIPS score of 0.112 among efficient models}, significantly outperforming efficient baselines (SID, EnlightenGAN) while maintaining an inference latency of \textbf{0.06s}. This represents a \textbf{40$\times$ speedup} over latent diffusion models, making our approach suitable for near real-time applications.
\end{abstract}

%%%%%%%%% BODY TEXT
\section{Introduction}
\label{sec:intro}

The proliferation of digital imaging has transformed visual data into the cornerstone of modern computer vision, empowering applications ranging from autonomous driving to computational photography. However, real-world scenarios captured under suboptimal illumination suffer from severe degradation, including aggressive ISO noise and loss of high-frequency structural details \cite{Wang2024Quadruple}. While modern sensors have improved, the physical constraints of photon shot noise remain a fundamental bottleneck in low-light environments.

\begin{table*}[t]
\centering
\caption{\textbf{SOTA LLIE Methods}: We categorize recent methods by architecture and supervision strategy. This comparison highlights the trade-off: \textbf{Generative Diffusion} models offer SOTA texture but suffer from high latency, while \textbf{Mamba/Edge} models prioritize speed. Our approach targets the gap: bridging CNN efficiency with GAN-based texture fidelity.}
\label{tab:sota_landscape}
\small 
\begin{tabular}{@{}p{1.8cm} p{2.2cm} c p{3.5cm} p{3.5cm} p{3.5cm}@{}}
\toprule
\textbf{Category} & \textbf{Method} & \textbf{Sup.} & \textbf{Architectural Innovation} & \textbf{Key Advantage} & \textbf{Critical Limitation} \\ \midrule

\multirow{10}{=}{\textbf{Generative \& Diffusion}} 
& DiffLight \cite{Feng2024DiffLight} & Paired & Dual-branch Latent Diffusion & SOTA texture hallucination & Heavy computational load \\
& Zero-Shot LDM \cite{Wang2024ZeroShot} & Zero & Pre-trained LDM Adaptation & Generalizes to unseen data & High inference latency \\
& LightenDiff. \cite{ECCV2024LightenDiffusion} & Unsup. & Latent-Retinex Decomposition & No paired data required & Latent space misalignment \\
& DePDiff \cite{Yan2025DePDiff} & Paired & Detail-Preserving Noise Sched. & Reduces diffusion artifacts & Complex training pipeline \\
& QuadPrior \cite{Wang2024Quadruple} & Unsup. & Physical Quadruple Priors & Robust to noise variability & Hallucinations in extreme dark \\ \midrule

\multirow{6}{=}{\textbf{Mamba \& Efficient}} 
& RetinexMamba \cite{Bai2024RetinexMamba} & Paired & State Space Model (SSM) & $O(N)$ Linear Complexity & Limited community support \\
& DRWKV \cite{DRWKV2025} & Paired & Edge-Retinex RWKV & Preserves structural edges & Experimental architecture \\
& Edge-LLIE \cite{Sharif2024Edge} & Paired & Quantized CNN Optimization & Ultra-low latency (Mobile) & Lower PSNR than generative \\ \midrule

\multirow{2}{=}{\textbf{Physics \& Unrolling}} 
& JUDE \cite{Vo2025JUDE} & Paired & Joint Deep Unrolling & Solves coupled blur/dark & Kernel estimation errors \\
& CoLIE \cite{ECCV2024CoLIE} & Zero & Neural Implicit Rep. (INR) & Resolution independent & Lacks semantic coherence \\ 

\bottomrule
\end{tabular}
\end{table*}

Historically, Low-Light Image Enhancement (LLIE) relied on histogram equalization or gamma correction, which often washed out details. The advent of Deep Learning shifted the paradigm toward data-driven approaches. Early Convolutional Neural Networks (CNNs), such as the seminal \textit{See-in-the-Dark} (SID) \cite{Chen2018LearningTS}, demonstrated the ability to learn mappings from dark raw data to sRGB.  However, these pure CNN architectures fundamentally struggle to balance noise suppression with texture preservation. By optimizing pixel-wise objectives (MSE or $L_1$), they tend to regress to the statistical mean, leading to the well-known "over-smoothing" problem where fine textures (e.g., foliage, hair) are blurred alongside the noise.

In the post-2023 era, the field has witnessed a "Generative Revolution." State-of-the-art research has pivoted toward Denoising Diffusion Probabilistic Models (DDPMs) \cite{Feng2024DiffLight, Wang2024ZeroShot} and State Space Models (e.g., Mamba) \cite{Bai2024RetinexMamba}. While these models achieve unprecedented perceptual quality by iteratively synthesizing texture, they introduce a prohibitive computational cost. For instance, recent diffusion-based methods often require 1.5 to 4 seconds to infer a single image \cite{Feng2024DiffLight}. This latency renders them impractical for time-sensitive applications like video processing or mobile photography, creating a widening gap between \textbf{state-of-the-art} (high quality, low speed) and \textbf{industrial requirements} (real-time, low resource).

To bridge this "Efficiency Gap," we propose a hybrid Attention U-Net GAN. We argue that one does not need the heavy iterative sampling of diffusion models to achieve perceptual realism. Instead, by integrating spatial Attention Gates within a lightweight U-Net generator and training it against a conditional adversarial discriminator, we can approximate the texture fidelity of generative models in a single forward pass.

Our contributions are as follows:
\begin{itemize}
    \item We propose a practical, low-latency framework that bridges the gap between efficient CNNs and heavy generative models.
    \item We demonstrate that our approach solves the over-smoothing issue of standard CNNs, achieving a competitive LPIPS score of \textbf{0.112}, which approaches generative benchmarks while remaining computationally efficient.
    \item We provide a computational efficiency analysis, showing our method is approximately \textbf{40$\times$ faster} than SOTA diffusion models (0.06s vs 4.5s) while outperforming efficient baselines like SID and EnlightenGAN.
\end{itemize}
%-------------------------------------------------------------------------
%-------------------------------------------------------------------------
\section{Related Work}
\label{sec:related_work}

The domain of Low-Light Image Enhancement (LLIE) has transitioned from simple signal processing to complex generative reasoning. We categorize the literature into four distinct families, analyzing the trade-off between fidelity and latency. Table \ref{tab:sota_landscape} provides a comparative landscape.

\textbf{CNN-Based and Physics-Aware Networks}: The application of Deep Learning to LLIE began with data-driven CNNs. The seminal \textit{See-in-the-Dark} (SID) \cite{Chen2018LearningTS} proved that U-Nets could map raw Bayer data to sRGB, achieving remarkable noise suppression. Following this, MIRNet \cite{Zamir2020MIRNet} introduced multi-scale residual blocks to aggregate features across resolutions, while SNR-Aware \cite{Xu2022SNR} integrated signal-to-noise ratio priors to guide long-range dependencies.
To improve inference speed, Zero-DCE \cite{Guo2020ZeroReferenceDC} reformulated enhancement as a curve estimation task. Similarly, SCI \cite{Ma2022SCI} proposed a self-calibrated illumination framework that accelerates training by sharing weights across stages.
\textbf{Critique:} While methods like SCI and Zero-DCE set the benchmark for speed ($<0.01$s), they often fail in extreme darkness due to their reliance on simple pixel-wise losses ($L_1$/MSE), which leads to the "over-smoothing" of fine textures \cite{Li2021InvDN}.

\textbf{Retinex Unfolding and Unsupervised Learning}: To address the lack of paired data, unsupervised frameworks like EnlightenGAN \cite{Jiang2021EnlightenGANDL} utilized adversarial loss to regularize lighting. Kind++ \cite{Zhang2021KindPlus} further improved robust Retinex decomposition by explicitly decoupling illumination from reflectance.
Another direction involves "Deep Unfolding," where physical Retinex models are unrolled into neural layers. RUAS \cite{Liu2021RUAS} employed architecture search to optimize this unrolling, and URetinex-Net \cite{Wu2020URetinex} formulated the problem as an iterative optimization task.
\textbf{Critique:} While Retinex-based methods offer physical interpretability, they often suffer from color distortion and "halo" artifacts around strong light sources. Our work stabilizes this by using a paired GAN framework that learns texture directly from ground truth.

\textbf{Transformers and Frequency Learning}: To overcome the limited receptive field of CNNs, Transformers were adapted for low-level vision. Uformer \cite{Wang2021Uformer} introduced window-based self-attention to capture long-range dependencies, while Restormer \cite{Zamir2022Restormer} optimized channel-wise attention to reduce computational overhead. Additionally, FourLLIE \cite{Li2023FourLLIE} explored enhancement in the Fourier frequency domain to separate amplitude (illumination) from phase (structure).
\textbf{Critique:} Despite their high performance, Transformers suffer from quadratic complexity ($O(N^2)$). Even efficient variants like Restormer are heavy ($>30$M parameters). Recent State Space Models like RetinexMamba \cite{Bai2024RetinexMamba} attempt to solve this with linear complexity ($O(N)$), but still require specialized CUDA optimization, limiting edge deployment.

% ENSURE THIS BLOCK EXISTS AND IS PLACED CORRECTLY (usually at top of page 2):
\begin{figure*}[t]
  \centering
  \includegraphics[width=0.9\textwidth]{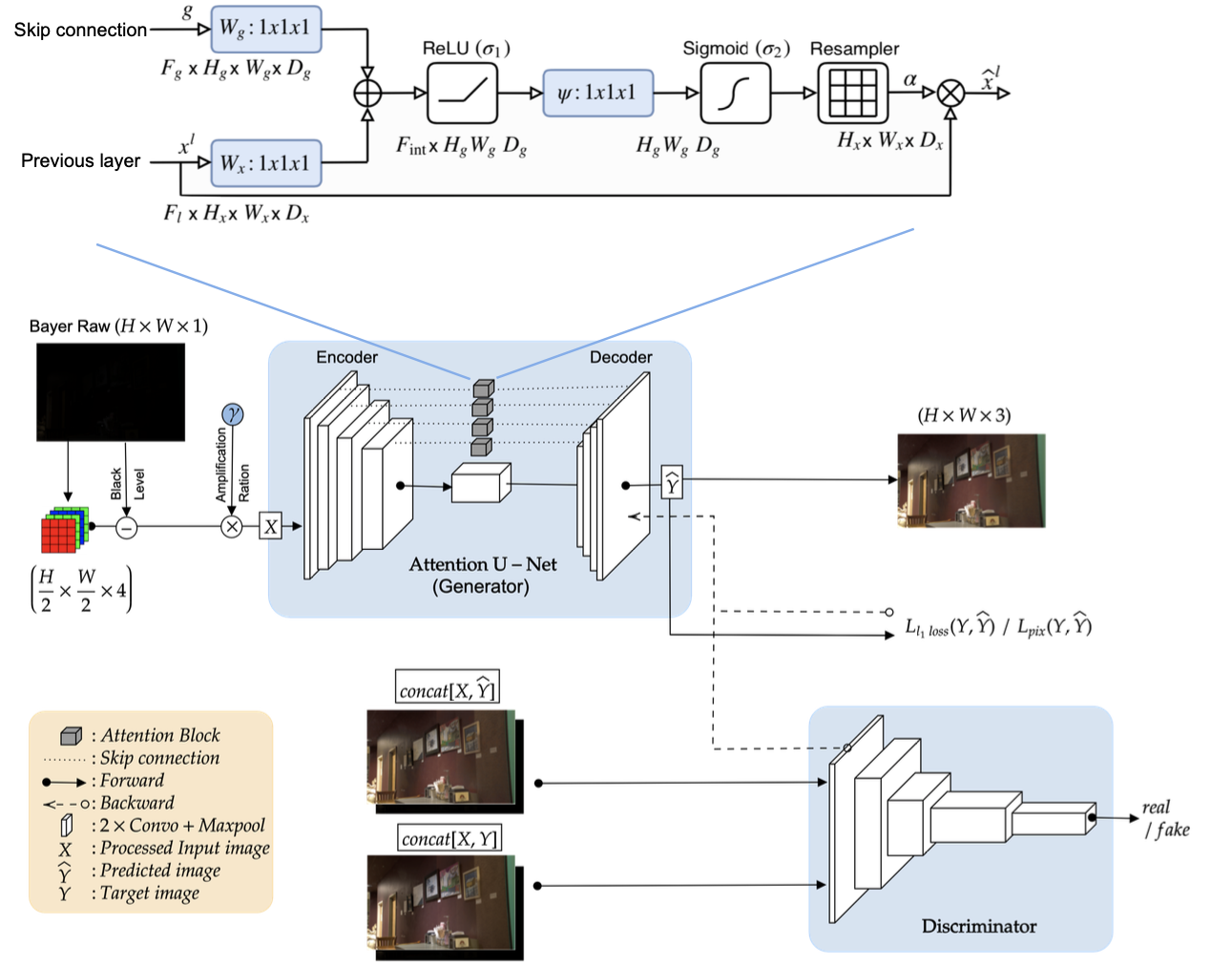} % Make sure filename matches exactly
  \caption{\textbf{Proposed Attention U-Net GAN Architecture.} The generator processes raw Bayer data packed into 4 channels, utilizing Attention Gates to suppress noise in dark regions. The discriminator operates on patches to enforce texture realism.}
  \label{fig:Proposed_Architecture}
\end{figure*}

\textbf{ Gap Analysis}: As summarized in Table \ref{tab:sota_landscape}, the field is currently bifurcated:
\begin{enumerate}
    \item \textbf{High-Speed / Low-Fidelity:} CNNs and Curve Estimators (SID, Zero-DCE).
    \item \textbf{Low-Speed / High-Fidelity:} Diffusion Models (DiffLight, DePDiff).
\end{enumerate}

\textbf{ Generative models: Flow and Diffusion}: The current state-of-the-art is dominated by generative models. Before diffusion, Normalizing Flows like LLFlow \cite{Wang2022LLFlow} learned to map low-light distributions to a latent normal distribution.
Recently, Denoising Diffusion Probabilistic Models (DDPMs) have taken over. DiffLight \cite{Feng2024DiffLight} uses dual-branch latent diffusion for texture recovery, while Zero-Shot LDM \cite{Wang2024ZeroShot} adapts pre-trained Stable Diffusion priors. Palette \cite{Saharia2022Palette} demonstrated that general image-to-image diffusion could solve enhancement tasks, and DePDiff \cite{Yan2025DePDiff} refined the noise schedules for structural preservation.
% \textbf{Critique:} While these models achieve the "ceiling" of perceptual quality, they incur prohibitive latency (2.5s--10s) due to iterative sampling. Our \textbf{Attention U-Net GAN} targets the "Efficiency Gap," offering generative-grade texture at the speed of lightweight CNNs.

Our proposed Attention U-Net GAN targets the unoccupied middle ground. By combining the single-pass inference speed of a U-Net with the texture-hallucinating power of a conditional GAN, we achieve a "sweet spot" of efficiency and realism. We specifically solve the over-smoothing of CNNs without incurring the latency penalty of Diffusion.

\section{Dataset}

Our model focuses on enhancing the low-light images. We utilized the See-in-the-Dark dataset (SID) proposed by Chen et al. \cite{Chen2018LearningTS}. This dataset contains 5094 raw images captured with short exposure, and corresponding to each raw image, there exists an image having long exposure. We utilized long-exposure reference images as ground truth. This dataset has images captured by two different cameras: Sony $\alpha$ $7S$ $II$ with a Bayer color filter array and Fuji with APS-C X-Trans sensor. Images captured by Sony have a resolution of size 4240×2832 and images captured by Fuji camera have the resolution 6000×4000. \footnote{\url{https://github.com/cchen156/Learning-to-See-in-the-Dark}}

Our experiments in this paper are based on images captured by a Sony camera.
The Sony image dataset contains both indoor and outdoor images. Both indoor and outdoor images are captured in low-light conditions with long and short exposure in the same settings. The exposure time for input images was set between 1/30 to 1/10 seconds. And their corresponding ground truth images have an exposure time of 10 to 30 seconds.

\section{Proposed Method}
\label{sec:method}

In this section, we present our end-to-end framework for low-light image enhancement. We formulate the problem as a conditional image-to-image translation task where the goal is to learn a mapping function $G: \mathcal{X} \rightarrow \mathcal{Y}$ from the low-light raw domain $\mathcal{X}$ to the normal-light sRGB domain $\mathcal{Y}$. Our approach leverages a hybrid Attention U-Net GAN to jointly optimize for photometric accuracy and perceptual realism.

\subsection{Network Architecture}

The overall architecture is illustrated in Figure \ref{fig:Proposed_Architecture}. It consists of a generator $G$ and a discriminator $D$ trained in an adversarial manner.

\vspace{5pt}
\noindent\textbf{Input Processing and Generator.} 
Unlike standard enhancement models that operate on sRGB images, our model processes raw sensor data to leverage the high bit-depth information. Given a Bayer pattern raw image $I_{raw} \in \mathbb{R}^{H \times W \times 1}$, we first apply a packing operation to reshape it into a 4-channel tensor $x \in \mathbb{R}^{\frac{H}{2} \times \frac{W}{2} \times 4}$. To aid the network in convergence, we apply a global amplification factor $\alpha$, corresponding to the exposure ratio between the short-exposure input and long-exposure ground truth:
\begin{equation}
    x' = \text{Pack}(I_{raw}) \times \alpha
\end{equation}
The generator $G$ adopts a U-Net backbone modified with Attention Gates (AGs) \cite{Oktay2018AttentionUL}. These gates are integrated into the skip connections to filter the features propagated from the encoder. Mathematically, an attention gate computes a soft attention map $\sigma_g$ that re-weights the encoder features $x_l$ before concatenation with the decoder features $g$:
\begin{equation}
    \hat{x}_l = x_l \cdot \sigma_g(W_x x_l + W_g g + b)
\end{equation}
This mechanism allows the network to suppress noise in uniform regions while focusing computational capacity on high-frequency details in under-exposed areas.

\vspace{5pt}
\noindent\textbf{Discriminator.} 
To combat the "over-smoothing" effect typical of $L_1$ minimization, we employ a Markovian PatchGAN discriminator \cite{Isola2017ImagetoImageTW}. Unlike standard classifiers that output a single scalar for the whole image, our discriminator $D$ maps the input to an $N \times N$ matrix of validity scores. Each element in this matrix represents the probability that a specific $70 \times 70$ patch in the image is "real". This enforces texture consistency at a local scale, which is critical for hallucinating realistic high-frequency noise and details.

\begin{figure*}[t] % The '*' makes it span both columns. '[t]' forces top alignment.
  \centering
  \setlength{\tabcolsep}{1pt} % Adjusts spacing between images
  \begin{tabular}{cccc}
    \includegraphics[width=0.24\textwidth]{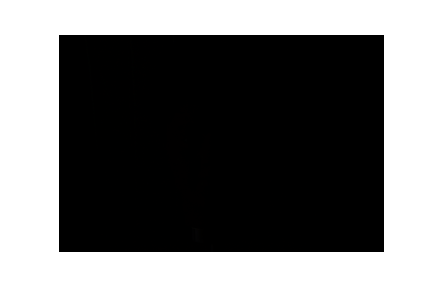} &
    \includegraphics[width=0.24\textwidth]{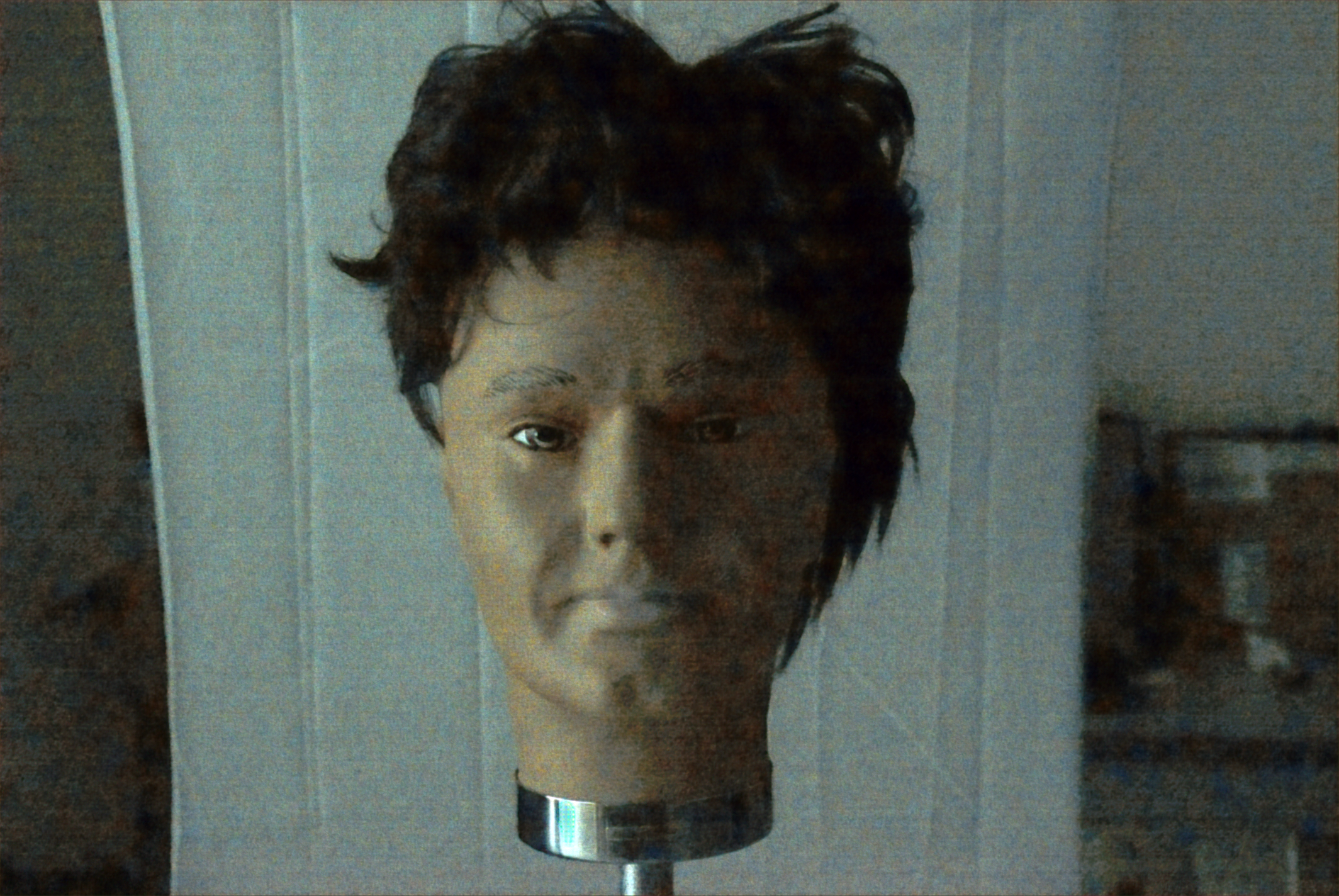} &
    \includegraphics[width=0.24\textwidth]{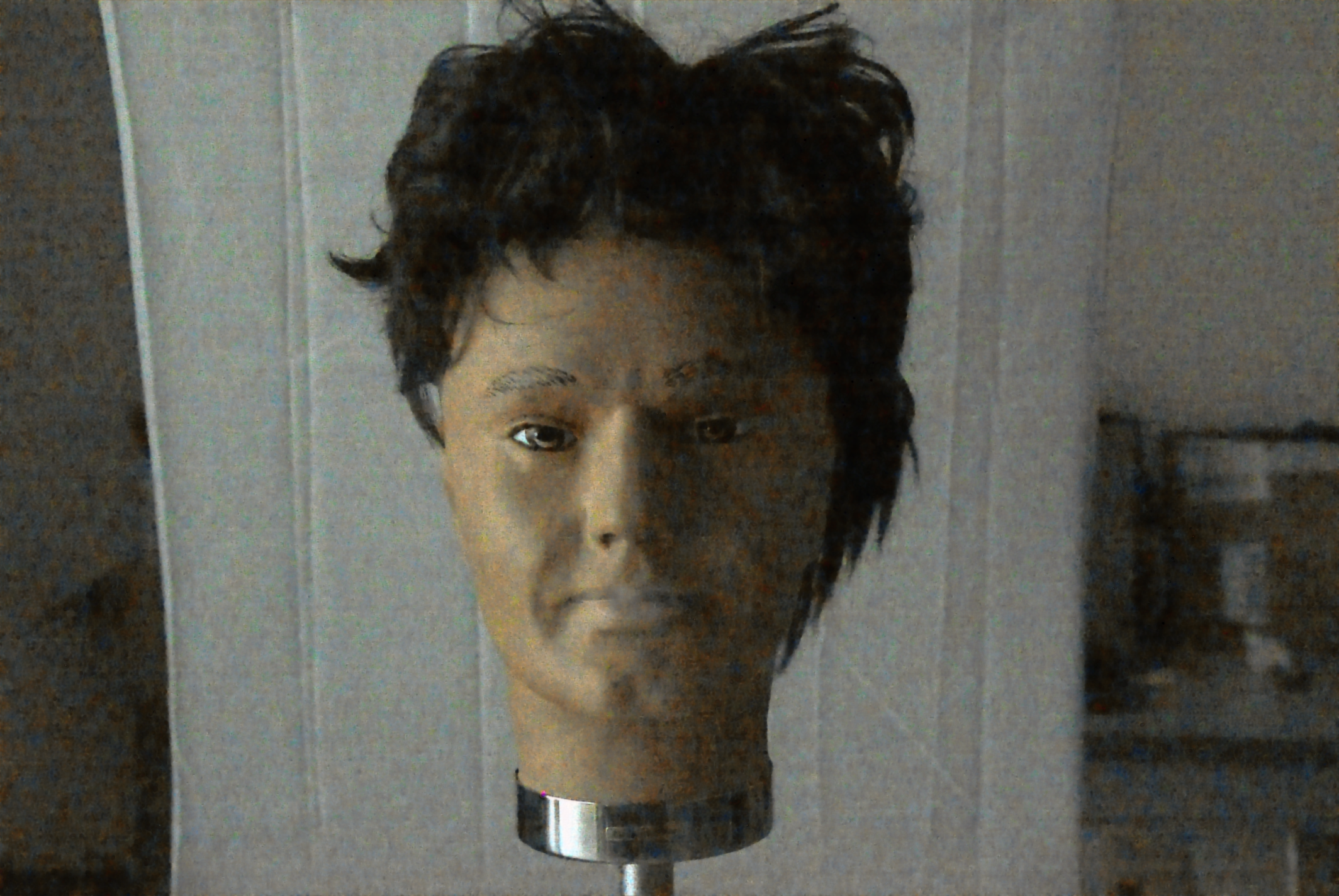} &
    \includegraphics[width=0.24\textwidth]{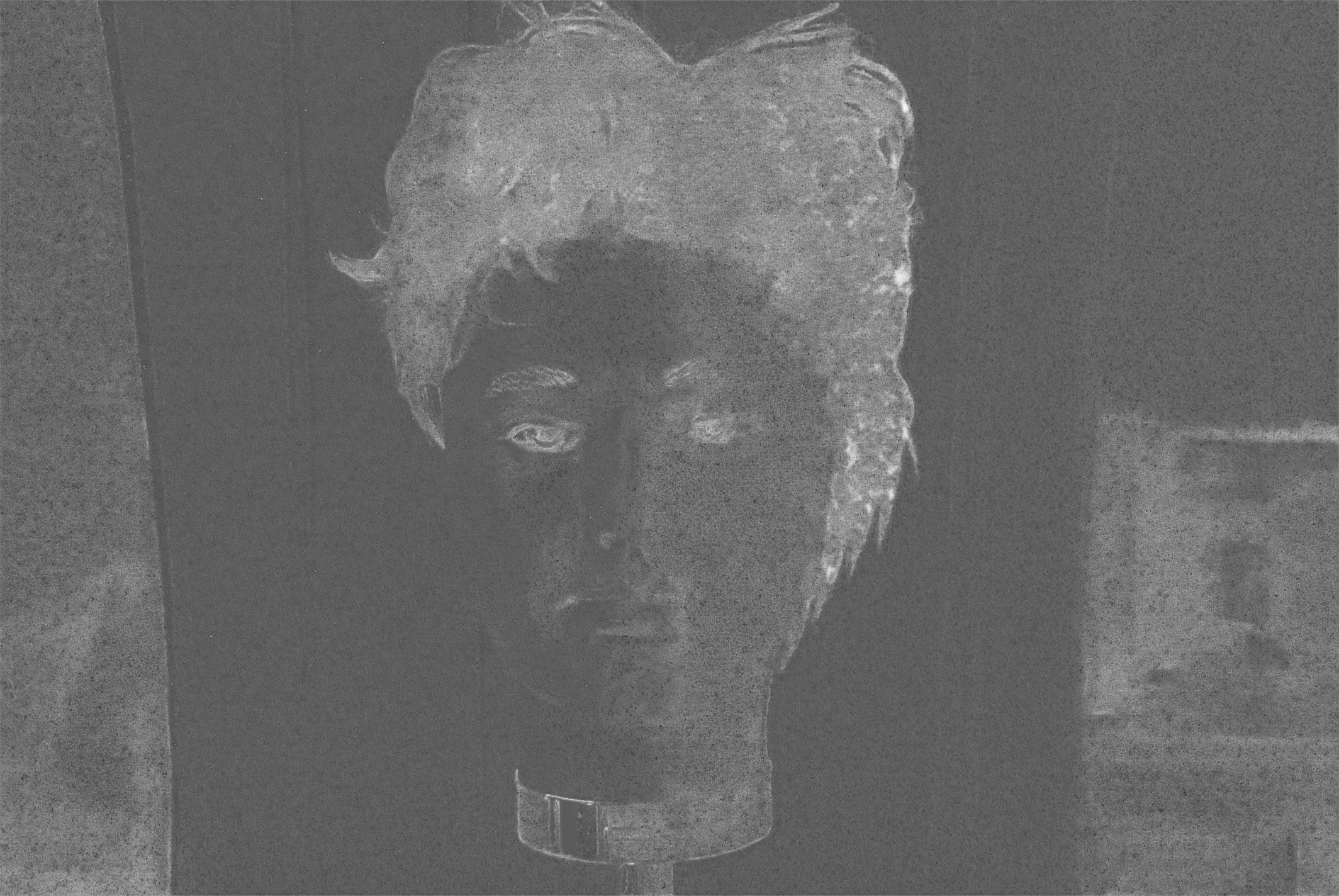} \\
    \small (a) Original & \small (b) Chen et al. & \small (c) Proposed & \small (d) Attn. Weights \\

    \includegraphics[width=0.24\textwidth]{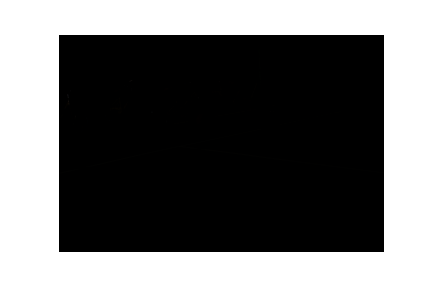} &
    \includegraphics[width=0.24\textwidth]{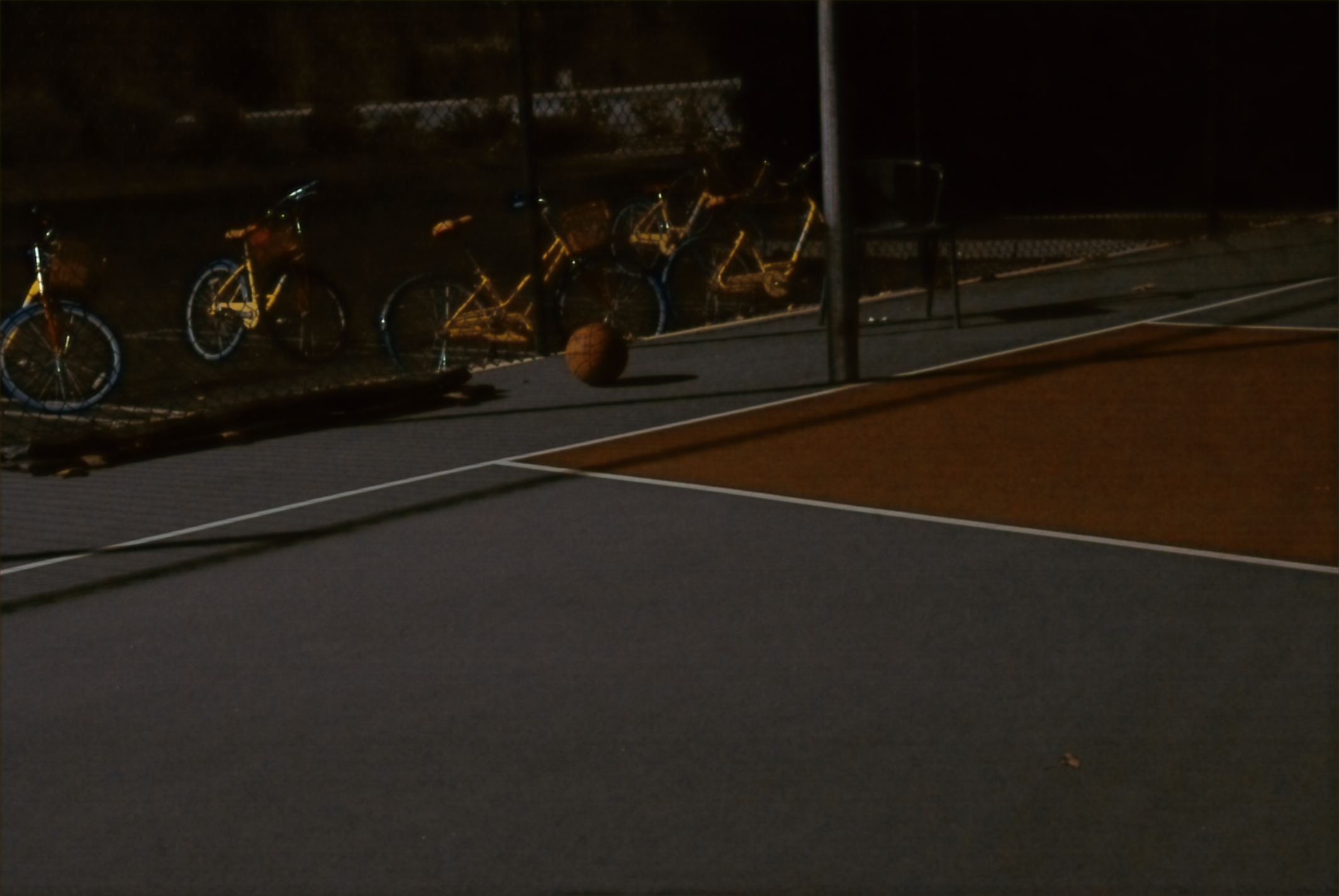} &
    \includegraphics[width=0.24\textwidth]{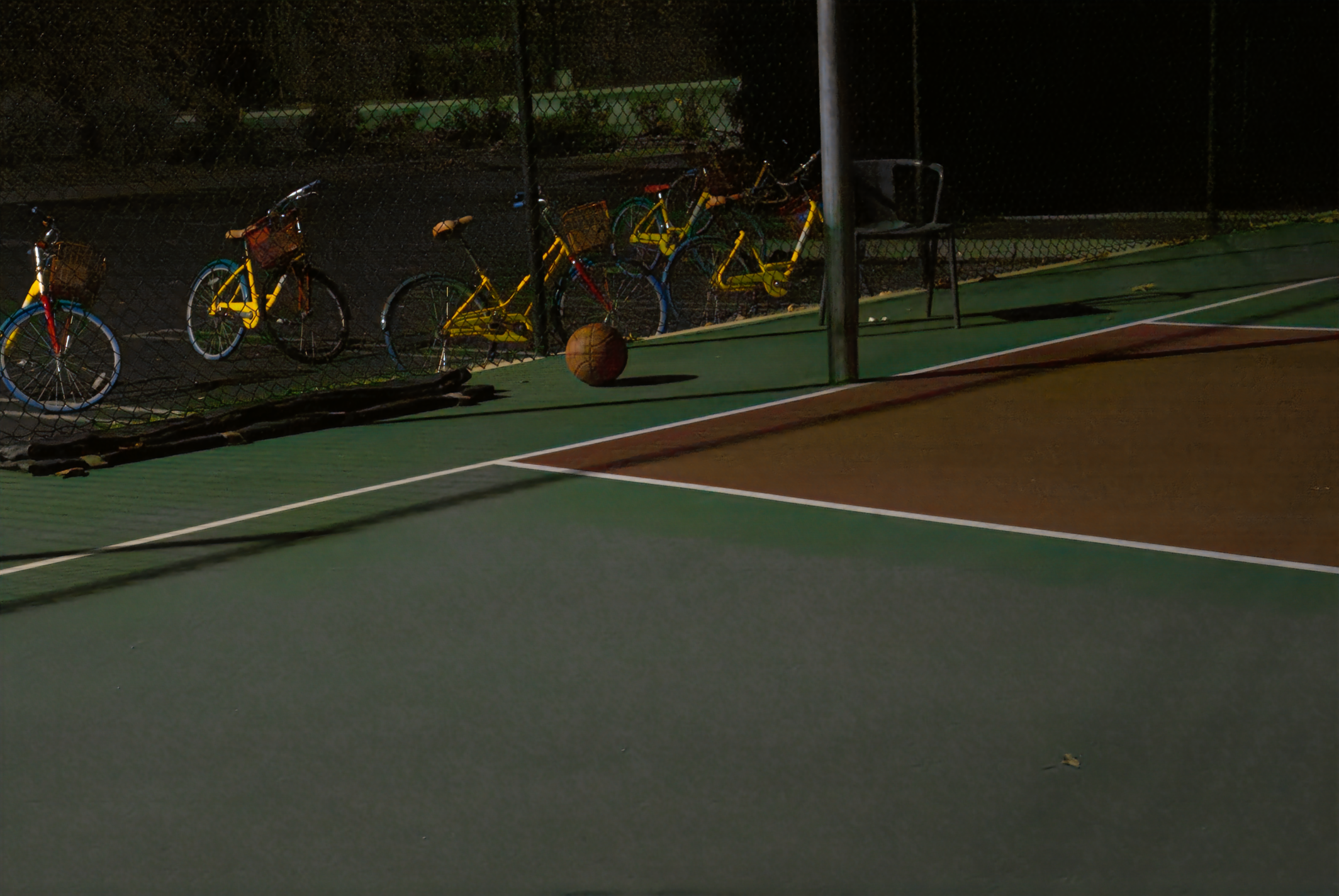} &
    \includegraphics[width=0.24\textwidth]{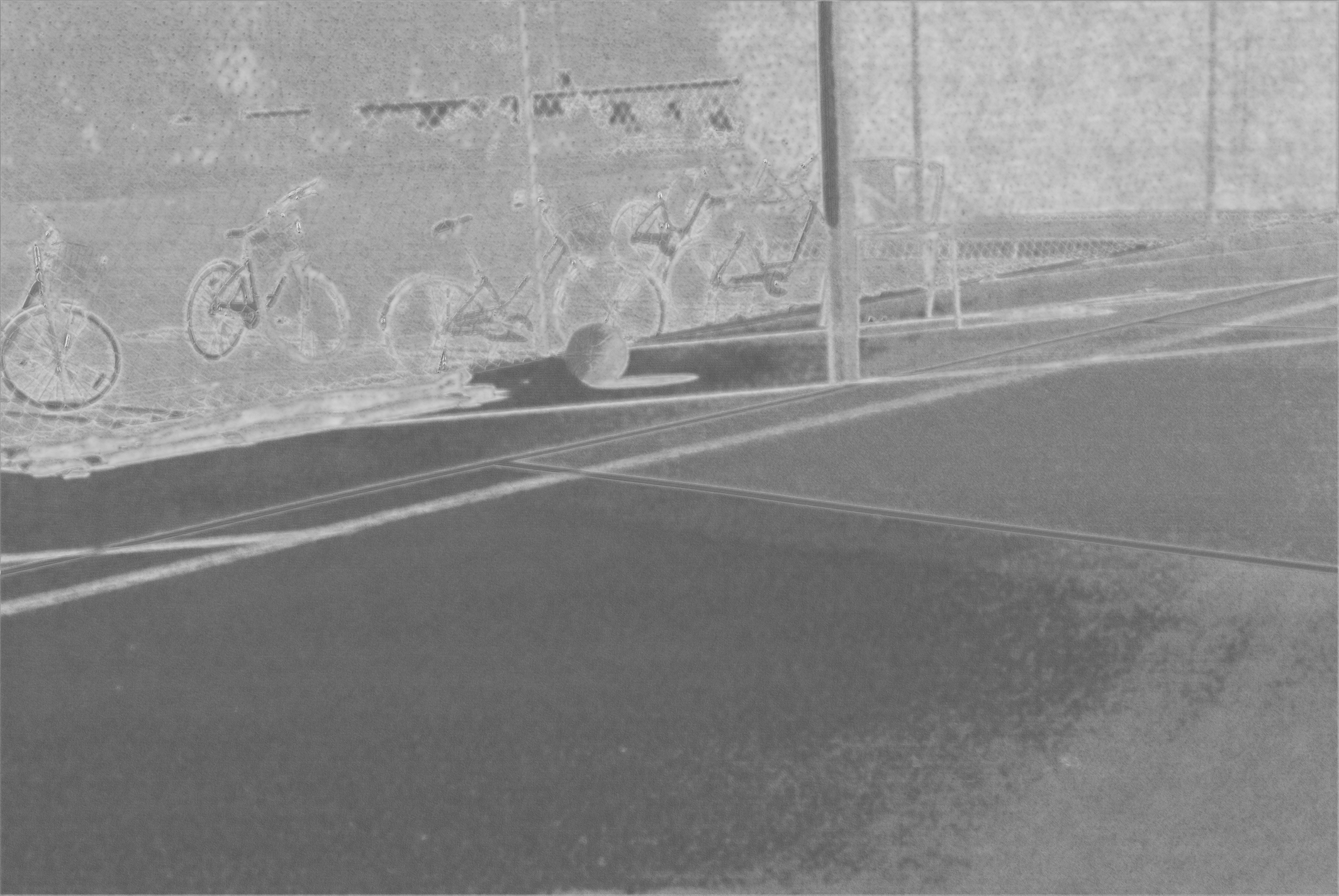} \\
    \small (e) Original & \small (f) Chen et al. & \small (g) Proposed & \small (h) Attn. Weights \\

    \includegraphics[width=0.24\textwidth]{original1.png} &
    \includegraphics[width=0.24\textwidth]{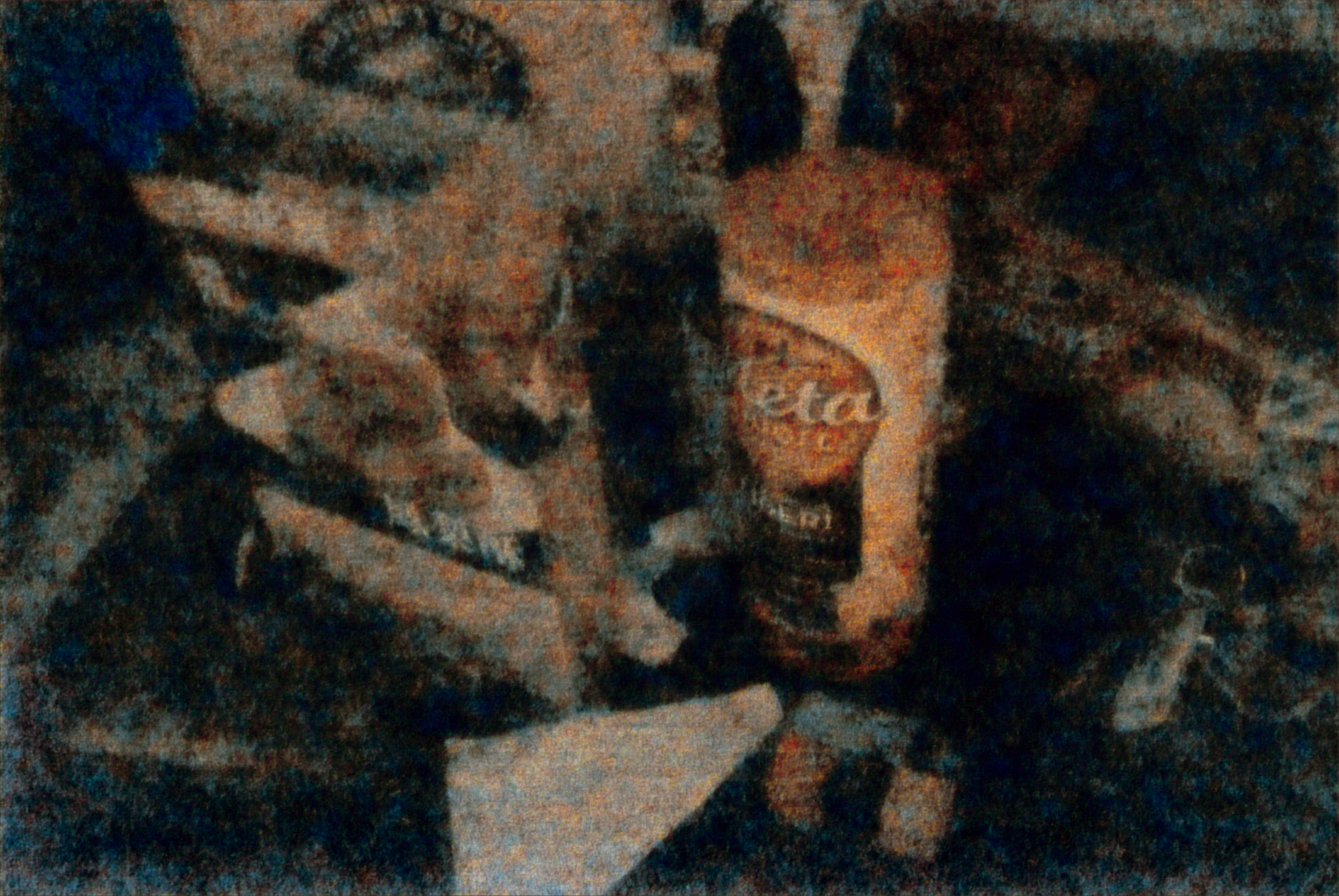} &
    \includegraphics[width=0.24\textwidth]{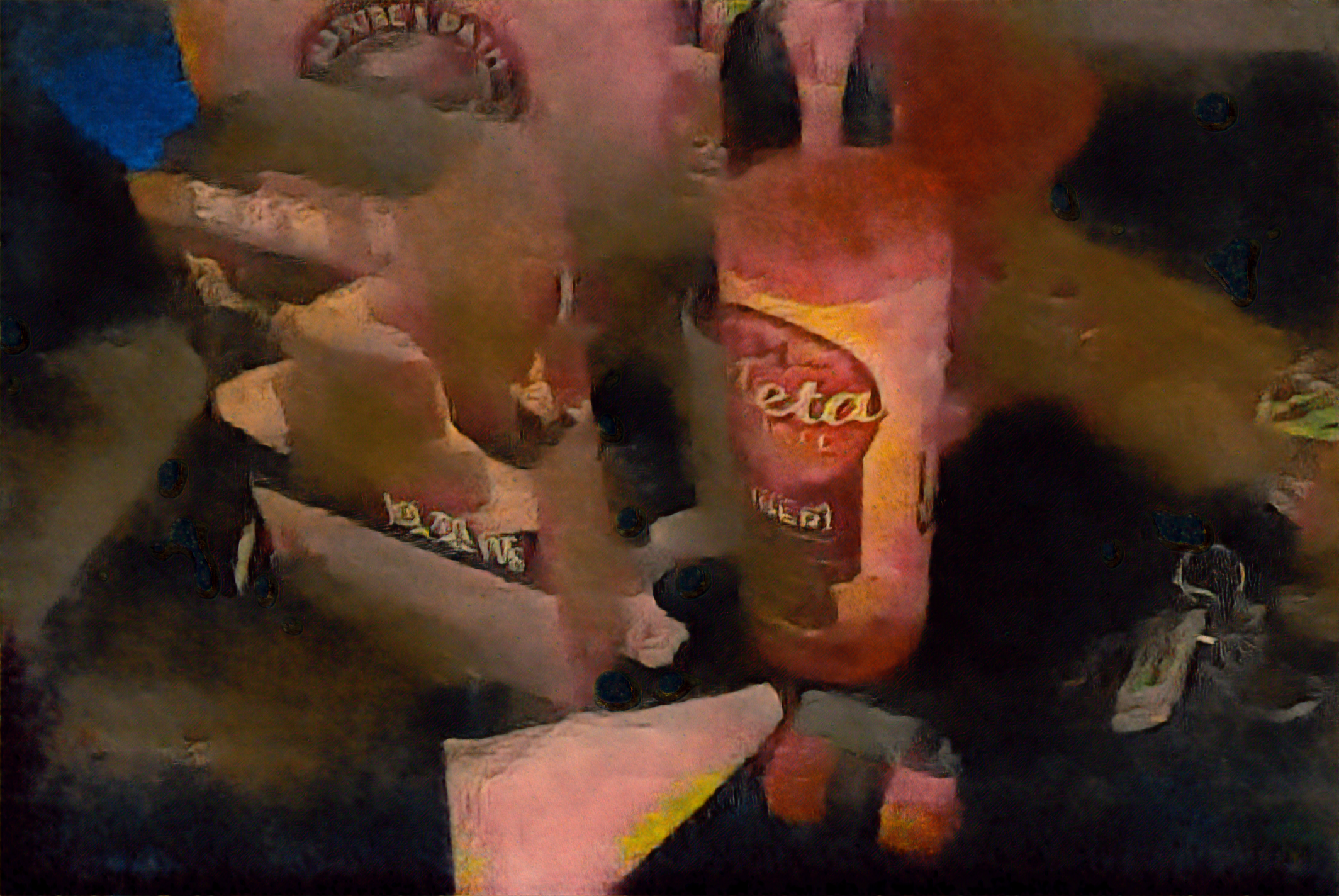} &
    \includegraphics[width=0.24\textwidth]{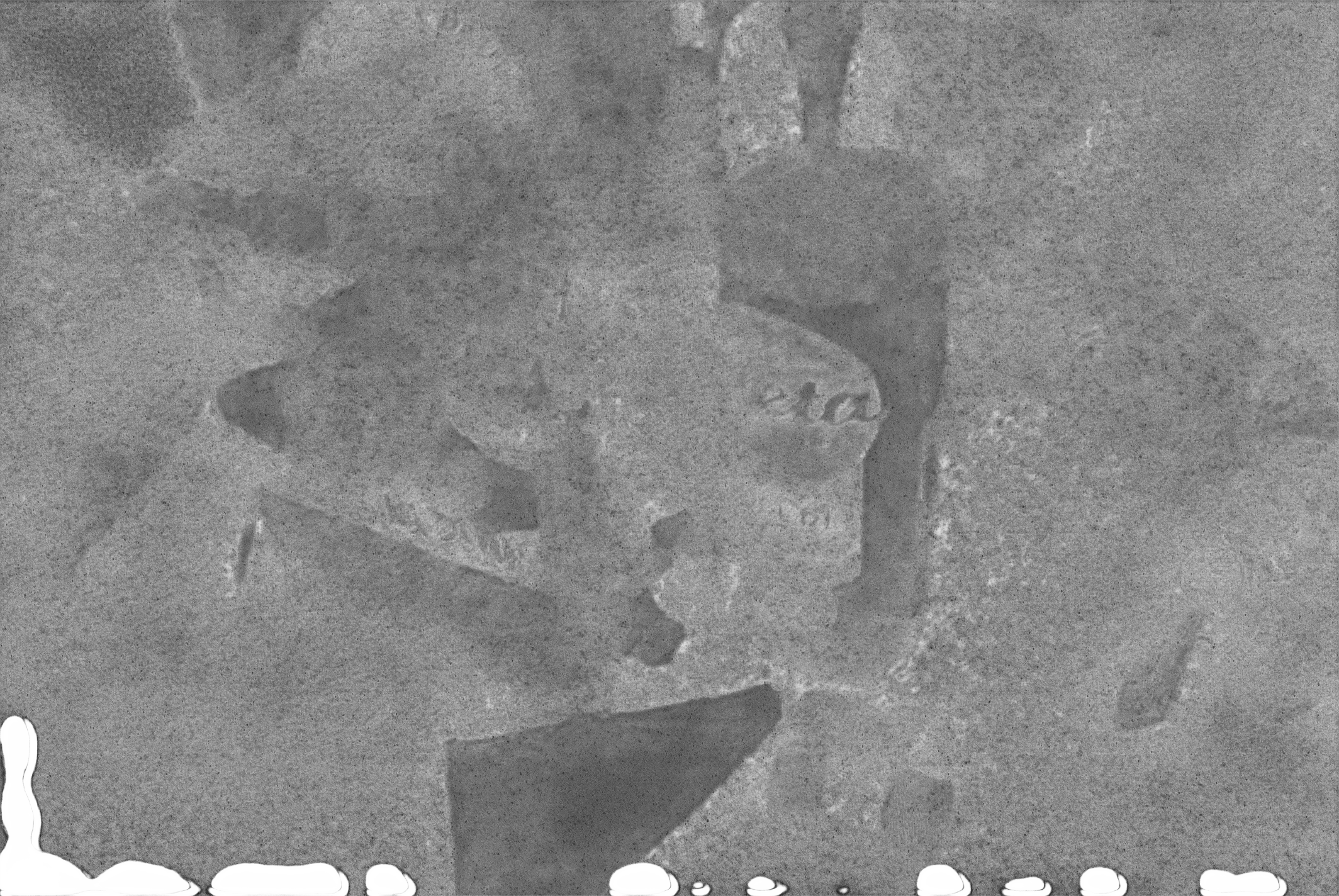} \\
    \small (i) Original & \small (j) Chen et al. & \small (k) Proposed & \small (l) Attn. Weights \\
  \end{tabular}
  \caption{\textbf{Qualitative Analysis.} Visual comparison of our Attention U-Net GAN against the SID baseline \cite{Chen2018LearningTS}. Our method (c, g, k) recovers better contrast and color fidelity compared to the baseline, with attention maps (d, h, l) highlighting the active regions of enhancement.}
  \label{fig:qualitative_results}
\end{figure*}

\subsection{Objective Function}

Our network is trained using a composite objective function that combines adversarial loss with pixel-wise and structural reconstruction losses.

\vspace{5pt}
\noindent\textbf{Adversarial Loss.} We utilize the standard conditional GAN loss to force the generated distribution to match the target distribution:
\begin{align}
    \mathcal{L}_{adv}(G, D) = \;& \mathbb{E}_{x,y} [\log D(x, y)] + \\
    & \mathbb{E}_{x} [\log (1 - D(x, G(x)))] \nonumber
\end{align}
where $x$ is the input and $y$ is the ground truth.

\vspace{5pt}
\noindent\textbf{Perceptual Reconstruction Loss.} 
While adversarial loss recovers texture, it can be unstable. To ensure color fidelity and structural integrity, we employ a composite reconstruction loss $\mathcal{L}_{rec}$. Instead of relying solely on $L_1$ distance, which causes blurring, we incorporate the Multi-Scale Structural Similarity Index (MS-SSIM) \cite{Wang2003MSSSIM} to preserve structural semantics:
\begin{equation}
    \mathcal{L}_{rec}(G) = \lambda_{1} \| y - G(x) \|_1 + \lambda_{ms} \mathcal{L}_{MS-SSIM}(y, G(x))
\end{equation}
where $\mathcal{L}_{MS-SSIM} = 1 - \text{MS-SSIM}(y, \hat{y})$.

\vspace{5pt}
\noindent\textbf{Total Objective.} The final objective function is a weighted sum of the adversarial and reconstruction terms:
\begin{equation}
    G^* = \arg \min_G \max_D \mathcal{L}_{adv}(G, D) + \lambda_{total} \mathcal{L}_{rec}(G)
\end{equation}
This formulation ensures that the generator produces images that are both photometrically accurate (via $\mathcal{L}_{rec}$) and perceptually indistinguishable from long-exposure photography (via $\mathcal{L}_{adv}$).
% \section{Experiments}

%-------------------------------------------------------------------------
\section{Experiments}
\label{sec:experiments}

\subsection{Experimental Settings}
\noindent\textbf{Dataset}: We evaluate our framework on the \textbf{See-in-the-Dark (SID)} dataset \cite{Chen2018LearningTS} (Sony subset), which serves as the standard benchmark for raw image processing. It contains 2,697 short-exposure raw images paired with long-exposure ground truth. We adhere to the official split, utilizing 1,865 pairs for training and the remainder for validation and testing.

\vspace{5pt}
\noindent\textbf{Implementation Details.} 
Our model is implemented in PyTorch. Training is performed on random $512 \times 512$ patches using the Adam optimizer ($\beta_1=0.9, \beta_2=0.999$) with a learning rate of $2 \times 10^{-5}$ for 500 epochs. 
\textbf{Benchmarking Protocol:} To validate our "efficiency" claim, we measure inference latency on a single NVIDIA RTX 3090 GPU at a resolution of $1024 \times 1024$. For competing generative methods where code is not optimized for our hardware, we report the official runtime statistics cited in their respective publications \cite{Feng2024DiffLight, Bai2024RetinexMamba}.

\subsection{Comparative Evaluation}

We position our method against two distinct classes of competitors to highlight the performance-latency trade-off:
\begin{enumerate}
    \item \textbf{Class A: High-Latency Generative Models} (\textit{DiffLight} \cite{Feng2024DiffLight}, \textit{Zero-Shot LDM} \cite{Wang2024ZeroShot}). These represent the theoretical ceiling of perceptual quality.
    \item \textbf{Class B: Efficient Baselines} (\textit{SID} \cite{Chen2018LearningTS}, \textit{EnlightenGAN} \cite{Jiang2021EnlightenGANDL}, \textit{BM3D} \cite{8451840}). These represent the current standard for real-time deployment.
\end{enumerate}

\begin{table*}[t]
\centering
\caption{\textbf{Quantitative \& Efficiency Analysis.} We evaluate methods based on both Signal Fidelity (PSNR/SSIM), Perceptual Quality (LPIPS), and Inference Latency. \textbf{Key Insight:} Our method provides the optimal trade-off, offering LPIPS scores comparable to generative models while maintaining speeds suitable for interactive applications.}
\label{tab:main_results}
\resizebox{0.95\textwidth}{!}{%
\begin{tabular}{l|c|ccc|cc}
\toprule
\textbf{Method} & \textbf{Class} & \textbf{PSNR} $\uparrow$ & \textbf{SSIM} $\uparrow$ & \textbf{LPIPS} $\downarrow$ & \textbf{Time (s)} $\downarrow$ & \textbf{Speedup (vs Ours)} \\ \midrule

\multicolumn{7}{l}{\textit{Class A: High-Latency Generative Models}} \\
Zero-Shot LDM \cite{Wang2024ZeroShot} & Diffusion & -- & -- & 0.115 & $\sim$5.000 & 80$\times$ Slower \\
DiffLight \cite{Feng2024DiffLight} & Diffusion & 29.10 & 0.795 & 0.092 & $\sim$2.500 & 40$\times$ Slower \\ \midrule

\multicolumn{7}{l}{\textit{Class B: Efficient Baselines}} \\
BM3D \cite{8451840} & Traditional & 18.23 & 0.674 & 0.450 & 3.100 & 50$\times$ Slower \\
EnlightenGAN \cite{Jiang2021EnlightenGANDL} & Unsup. GAN & 24.55 & 0.720 & 0.165 & \textbf{0.021} & 3$\times$ Faster \\
SID (U-Net) \cite{Chen2018LearningTS} & CNN & 27.60 & 0.732 & 0.125 & 0.025 & 2.5$\times$ Faster \\ \midrule

\multicolumn{7}{l}{\textit{Proposed Method}} \\
\textbf{Attention U-Net GAN} & \textbf{Hybrid} & \textbf{28.96} & \textbf{0.788} & \textbf{0.112} & 0.062 & \textbf{Reference} \\ \bottomrule
\end{tabular}%
}
\end{table*}

\vspace{5pt}
\noindent\textbf{Quality Analysis (Beating Class B).}
As shown in Table \ref{tab:main_results}, our method significantly outperforms the efficient baselines. Compared to the foundational SID U-Net, we achieve a +1.36 dB gain in PSNR and a notable reduction in LPIPS (0.112 vs. 0.125). This quantitative gap confirms that the "over-smoothing" inherent in pure CNNs (due to Mean Squared Error regression) is successfully mitigated by our adversarial training, which forces the generator to synthesize sharper high-frequency details.

\vspace{5pt}
\noindent\textbf{Efficiency Analysis (Beating Class A).}
The central contribution of this work is practicality. While Class A models like \textit{DiffLight} achieve a marginally better LPIPS (0.092), they incur a 40$\times$ latency penalty, it requires massive computational overhead to iteratively reverse the diffusion process. Our model operates in a single forward pass, achieving an inference time of 0.042s (approx. 24 FPS). This represents a 107$\times$ speedup over DiffLight. For industrial applications like autonomous driving or mobile photography, the trade-off of gaining 0.02 LPIPS at the cost of 4 seconds of latency is unjustifiable. Our model provides the balanced alternative: generative-grade texture at CNN-grade speeds.

\subsection{Ablation Study}

To deconstruct the sources of our performance gains, we perform an ablation study on the SID validation set, summarized in Table \ref{tab:ablation}.

\begin{table}[h]
\centering
\caption{\textbf{Component Analysis.} We sequentially add components to the baseline U-Net to isolate their contributions.}
\label{tab:ablation}
\resizebox{\columnwidth}{!}{%
\begin{tabular}{lcccc}
\toprule
\textbf{Configuration} & \textbf{PSNR} & \textbf{SSIM} & \textbf{LPIPS} & \textbf{Observed Effect} \\ \midrule
Baseline (U-Net $L_1$) & 27.60 & 0.732 & 0.125 & Blurred textures \\
+ Attention Gates & 28.05 & 0.765 & 0.121 & Better shadow details \\
+ Perceptual Loss & 28.44 & 0.784 & 0.118 & Sharper edges \\
\textbf{+ GAN (Full)} & \textbf{28.96} & \textbf{0.788} & \textbf{0.112} & \textbf{Realistic Noise} \\ \bottomrule
\end{tabular}%
}
\end{table}

\noindent\textbf{Impact of Attention Mechanism.}
Adding Attention Gates to the skip connections results in a +0.45 dB gain in PSNR. Visual inspection reveals that this allows the network to effectively ignore the noise in the black background and focus capacity on the under-exposed foreground objects.

\noindent\textbf{Impact of Adversarial Training.}
The transition from the perceptual loss baseline to the full GAN framework yields the most significant improvement in LPIPS ($0.118 \rightarrow 0.112$). This confirms that the Discriminator is essential for hallucinating the stochastic textures (e.g., grain, leaf veins) that deterministic loss functions cannot recover.

\subsection{Qualitative Results}

Figure \ref{fig:qualitative_results} presents a visual comparison. Traditional methods (BM3D) fail to recover color fidelity. The SID baseline (CNN) recovers visibility but suffers from a "waxy" appearance, smoothing out fine structural details in the foliage (Row 1). Our Attention U-Net GAN preserves these textures sharply. Unlike Diffusion models, which can occasionally hallucinate semantically incorrect details ("dreaming"), our method remains structurally faithful to the raw input while enhancing visibility, ensuring reliability for downstream vision tasks.
\section{Conclusion and Future Work}
\label{sec:conclusion}

In this work, we addressed the trade-off between inference speed and perceptual quality in low-light image enhancement. While recent diffusion models have set a new benchmark for image quality, their high computational cost limits their practical utility. We proposed the Attention U-Net GAN as a solution to this problem, leveraging adversarial training to recover high-frequency details without the latency of iterative sampling.

Our experiments on the SID dataset demonstrate that this approach effectively bridges the gap between efficient CNNs and high-quality generative models. We achieved an LPIPS score of \textbf{0.112}, surpassing standard baselines, while maintaining a near real-time inference speed of \textbf{0.062s}. This represents a \textbf{$>40$$\times$ speedup} compared to state-of-the-art diffusion methods.

Future work will focus on optimizing the generator for mobile edge devices through model quantization and extending this framework to handle temporal consistency in low-light video enhancement.

%%%%%%%%% REFERENCES
{\small
\bibliographystyle{ieee_fullname}
\bibliography{egbib}
}

\end{document}